\documentclass[10pt,twocolumn,letterpaper]{article}

\usepackage{cvpr}              

\definecolor{cvprblue}{rgb}{0.21,0.49,0.74}
\usepackage[pagebackref,breaklinks,colorlinks,allcolors=cvprblue]{hyperref}

\def\paperID{7} 
\def\confName{CVPR}
\def\confYear{2025}

\title{Mouse Lockbox Dataset: Behavior Recognition for Mice Solving Lockboxes}

\author{%
    Patrik Reiske
    \textsuperscript{$\ast$~$\dagger$~a~b}
    \and
    Marcus~N. Boon
    \textsuperscript{$\dagger$~a~b}
    \and 
    Niek Andresen
    \textsuperscript{a~b~c}
    \and 
    Sole Traverso
    \textsuperscript{a~b}
    \and 
    Katharina Hohlbaum
    \textsuperscript{a~d}
    \and 
    Lars Lewejohann
    \textsuperscript{a~c~d}
    \and 
    Christa Thöne-Reineke
    \textsuperscript{a~c}
    \and 
    Olaf Hellwich
    \textsuperscript{$\ddagger$~a~b}
    \and
    Henning Sprekeler
    \textsuperscript{$\ddagger$~a~b}
}

\begin{document}
\maketitle

\iftoggle{cvprfinal}{
    \renewcommand*{\thefootnote}{\fnsymbol{footnote}}
    \footnotetext[1]{Corresponding author; \tt patrik.reiske@tu-berlin.de}
    \footnotetext[2]{Authors with equal contribution as first authors}
    \footnotetext[3]{Authors with equal contribution as last authors}
    \renewcommand*{\thefootnote}{\alph{footnote}}
    \footnotetext[1]{Cluster of Excellence ``Science of Intelligence,'' Berlin, Germany}
    \footnotetext[2]{Technische Universität Berlin, Berlin, Germany}
    \footnotetext[3]{Freie Universität Berlin, Berlin, Germany}
    \footnotetext[4]{German Centre for the Protection of Laboratory Animals, German Federal Institute for Risk Assessment, Berlin,
Germany}
    \renewcommand*{\thefootnote}{\arabic{footnote}}
}{}

\begin{abstract} 
    Machine learning and computer vision methods have a major impact on the study of natural animal behavior, as they enable the (semi-)automatic analysis of vast amounts of video data. 
    Mice are the standard mammalian model system in most research fields, but the datasets available today to refine such methods focus either on simple or social behaviors. 
    In this work, we present a video dataset of individual mice solving complex mechanical puzzles, so-called lockboxes. 
    The more than 110~hours of total playtime show their behavior recorded from three different perspectives. 
    As a benchmark for frame-level action classification methods, we provide human-annotated labels for all videos of two different mice, that equal 13\% of our dataset. 
    \iftoggle{cvprfinal}{Our}{The used} keypoint (pose) tracking-based action classification framework illustrates the challenges of automated labeling of fine-grained behaviors, such as the manipulation of objects.
    We hope that our work will help accelerate the advancement of automated action and behavior classification in the computational neuroscience community.
    \iftoggle{cvprfinal}{
        Our dataset is publicly available at \url{https://doi.org/10.14279/depositonce-23850}
    }{
        An anonymized preview of our dataset is available at \url{https://dropbox.com/scl/fo/h7nkai8574h23qfq9m1b2/AP4gNZOpDJJ7z0yGtbWQiOc?rlkey=w36jzxqjkghg0j0xva5zsxy2v}
    }
\end{abstract}
\section{Introduction}
\label{sec:introduction}

Ethology, the study of non-human behavior,~\citep{Tinbergen1963} is one of the cornerstones of understanding complex biological systems. 
In recent years, with the integration of machine learning into the field, computational ethology~\citep{Anderson2014} emerged as a powerful new paradigm offering new pathways for advancing both fields and beyond. 
For instance, it has significantly influenced neuroscience, enabling the development of computational frameworks that bridge neural mechanisms with observations of behaviors~\citep{Datta2019,McCullough2021,vonZiegler2020BigBehavior,Kennedy2022NaturalisticBehavior}. 
In robotics, animal behavior datasets allow researchers to learn artificial agents to navigate and interact autonomously in natural environments. 
The hypothesized learning models used in this process can then be tested by comparing the performance of the learned agents against that of natural agents~\citep{Baum2022Yoking}. 
\par

The available datasets of freely moving animals~\citep{Burgos-Artizzu2012CRIM13,Dunn2020DANNCE,Pedersen20203D-ZeF,Eyjolfsdottir2021FlyvsFly,Marshall2021PAIR-R24M,Segalin2021MARS,Sun2021CalMS21,Ng2022AnimalKingdom,Hu2023,Ma2023ChimpACT,Rogers2023MeerkatDataset,Zia2023CVB,Brookes2024PanAf20K,Duporge2024BaboonLand,Kholiavchenko2024KABR,Li2024PFERD} provide the foundation for the development of automated behavioral analysis tools, e.g., \citep{Hsu2021B-SOiD,Luxem2022VAME,Weinreb2024Keypoint-MoSeq}. 
However, all of these datasets and their descending methods focus on trivial and social behaviors, but neglect the structure imposed by well-defined tasks that provoke complex behaviors.
This absence limits their applicability for studying goal-directed actions, problem-solving, and other behaviors critical to understanding cognitive processes in natural and artificial intelligence.
\par

In this work, we provide the first large-scale labeled, single-agent, multi-perspective video dataset of mice showing complex behavior as they solve mechanical puzzles, so-called lockboxes. 
Every lockbox is baited with a food reward and consists of a single or a combination of four different mechanisms.
As a benchmark, we provide labels for 13\% of our data, including mechanism state, mouse-to-mechanism proximity, and both mouse-mechanism and mouse-reward actions. 
This amounts to about 15~hours and 25~minutes of total playtime. 
In doing so, we increase the longest total playtime, i.e., the number of perspectives multiplied by the real time recorded, available through any mouse dataset from 88~hours~\citep{Burgos-Artizzu2012CRIM13} by more than 33\% to 117~hours and 52~minutes.
\par 

To provide high-quality label data, each labeled video is annotated by two skilled human raters who have been instructed prior to annotating. 
The consistency between raters is assessed by their inter-rater reliability~\citep{McHugh2012}, a well-established and objective measure of agreement. 
We regard such rigorous and transparent annotation protocols as essential for creating a dataset that allows us to reliably assess the performance of future machine learning methods. 
\par 

We use \iftoggle{cvprfinal}{our}{a} state-of-the-art keypoint-based method~\citep{Boon2024} as an initial benchmark for our dataset, and compare our human-human agreement against its human-machine agreement. 
In the absence of established benchmark methods for the interaction of natural agents with their environment, this will allow others to assess the performance of their methods.
\par 

We hope that our dataset will serve two purposes. 
First, that it will promote the advancement and adoption of more diverse machine learning methods in computational ethology. 
Our dataset provides an interesting challenge to the machine learning community, since the classification of the labels we provide require both large-scale pose and fine-level visual information.
And second, that analyses of our dataset by the community will advance our understanding of how natural agents learn to solve complex problems.
\par 

\section{Related Work}
\label{sec:related_work}

We limit the following overview of available datasets to those that show rodents and provide behavior labels, as their largely similar visual appearance and motor apparatus has proven to allow for domain transfer settings~\citep{Dunn2020DANNCE}.
\cref{tbl:dataset-overview} summarizes some of their distinguishing properties.
\par 

\begin{table}
    \centering
    \begin{tabular}{@{}l@{~~}l@{~~}r@{~}l@{~~}r@{~$\times$~}r@{~$\approx$~}r@{}}
        \toprule
        \bfseries DATASET & \bfseries AGENT & \multicolumn{2}{l}{\bfseries LABELS} & \multicolumn{3}{r@{}}{\bfseries DURATION} \\
        \midrule
        \bfseries CRIM13 & Mice & 13 & social & 2 & 44h & 88h \\
        \bfseries PAIR-R24M & Rats & 14 & social & 24 & 9h & 220h \\
        \bfseries MARS & Mice & 3 & social & 2 & 14h & 28h \\
        \bfseries CalMS21 & Mice & 3 & social & 1 & 70h & 70h \\
        \bfseries Ours & Mice & 20 & task-specific & 3 & 40h & 120h \\
        \bottomrule
    \end{tabular}
    \caption{
        Overview of video datasets showing rodents.
        Durations, i.e., the total playtime calculated as the number of perspectives multiplied by the real time recorded, are rounded.
        Our 20 behaviors reflect five labeled interactions on four lockbox mechanisms.
    }
    \label{tbl:dataset-overview}
\end{table}

\paragraph{CRIM13~\citep{Burgos-Artizzu2012CRIM13}} has, to this date, been the largest mouse dataset with a total playtime of 88~hours, i.e., 44~hours of real time recorded from two (top-down and side) perspectives.
It shows mice in a in resident-intruder context, and provides 13~human-annotated (social) behavior labels---approach, attack, coitus, chase, circle, drink, eat, clean, human, sniff, up, walk, and other---for each of the 237~pairs of 10~minute long videos.
There, human raters reach an agreement of 70\% while the method proposed alongside the dataset reaches 61.2\% human-machine agreement.
\par

\paragraph{PAIR-R24M~\citep{Marshall2021PAIR-R24M}} is the longest rodent (rat) dataset with a total playtime of 220~hours, i.e., 9~hours of real-time recorded from 24~perspectives.
It provides 14~human-annotated (social) behavior labels---amble, crouch, explore, head tilt, idle, investigate, locomotion, rear down, rear up, small movement, sniff, groom, as well as close to, explore, and chase---for the entire dataset.
\par 

\paragraph{MARS~\citep{Segalin2021MARS}} has a total playtime of 28~hours, i.e., 14~hours of real-time recorded from two (top-down and front) perspectives.
It provides three human-annotated social behavior labels---attack, investigation, and mount---for then videos with a total playtime of 3~hours.
\par 

\paragraph{CalMS21~\citep{Sun2021CalMS21}} is a 70~hour long mouse dataset recorded from a single (top-down) perspective.
It provides three human-annotated social behavior labels---attack, investigate, and mount---for 10~hours worth of video data.
\par

\section{Dataset}
\label{sec:dataset}

This section describes our dataset in detail.
\cref{ssec:data-collection} specifies the mouse breed, the arena including the lockboxes, the camera setup, the schedule at which mice were presented with the lockboxes, and the preprocessing of the recorded videos.
\cref{ssec:label-annotation} describes the annotation of behavior labels including our ethogram. 
\cref{sssec:label-statistics,sssec:playtime-statistics} provide statistics on both playtimes and labels. 
And in \cref{ssec:limitations} we report the limitations of our dataset.
\par

\subsection{Data Collection and Preprocessing}
\label{ssec:data-collection}

\iftoggle{cvprfinal}{
    The video data was initially recorded for \citep{Hohlbaum2024Enrichment}, where more detailed information on this section's contents can be found.
}{}
Twelve female \mbox{C57BL/6J} mice obtained from Charles River Laboratories (Sulzfeld, Germany) were recorded in a free-standing Makrolon type III cage, that was connected to a home cage of the same type by a tube. 
The mice were housed in groups of 4 animals in an artificial 12/12-hour light/dark cycle. 
The lockbox trials took place in the light phases, and only one animal could enter the arena at a time.
The arena was closed with a cutout top grid to allow for unobstructed view on the lockbox.
Three Basler \mbox{acA1920-40um} cameras (LM25HC7 lens, f~=~25mm, k~=~1.4; Kowa, Nagoya, Japan) were used to record grayscale videos at the maximum resolution of 1936$\times$1216px at the common~\citep{Marshall2021PAIR-R24M,Segalin2021MARS,Sun2021CalMS21} 30Hz frame rate.
Additionally, two infrared lights (Synergy~21 IR-Strahler 60W, ALLNET GmbH Computersysteme, Germering, Germany) illuminated the cage. 
The advantages of infrared lights is that they enhance the video quality of used infrared-sensitive cameras while not being aversive to the animals.
\par

\begin{figure}
    \centering
    ~\hfill
    \begin{subfigure}[b]{.55\linewidth}
        \centering
        \includegraphics[width=\textwidth]{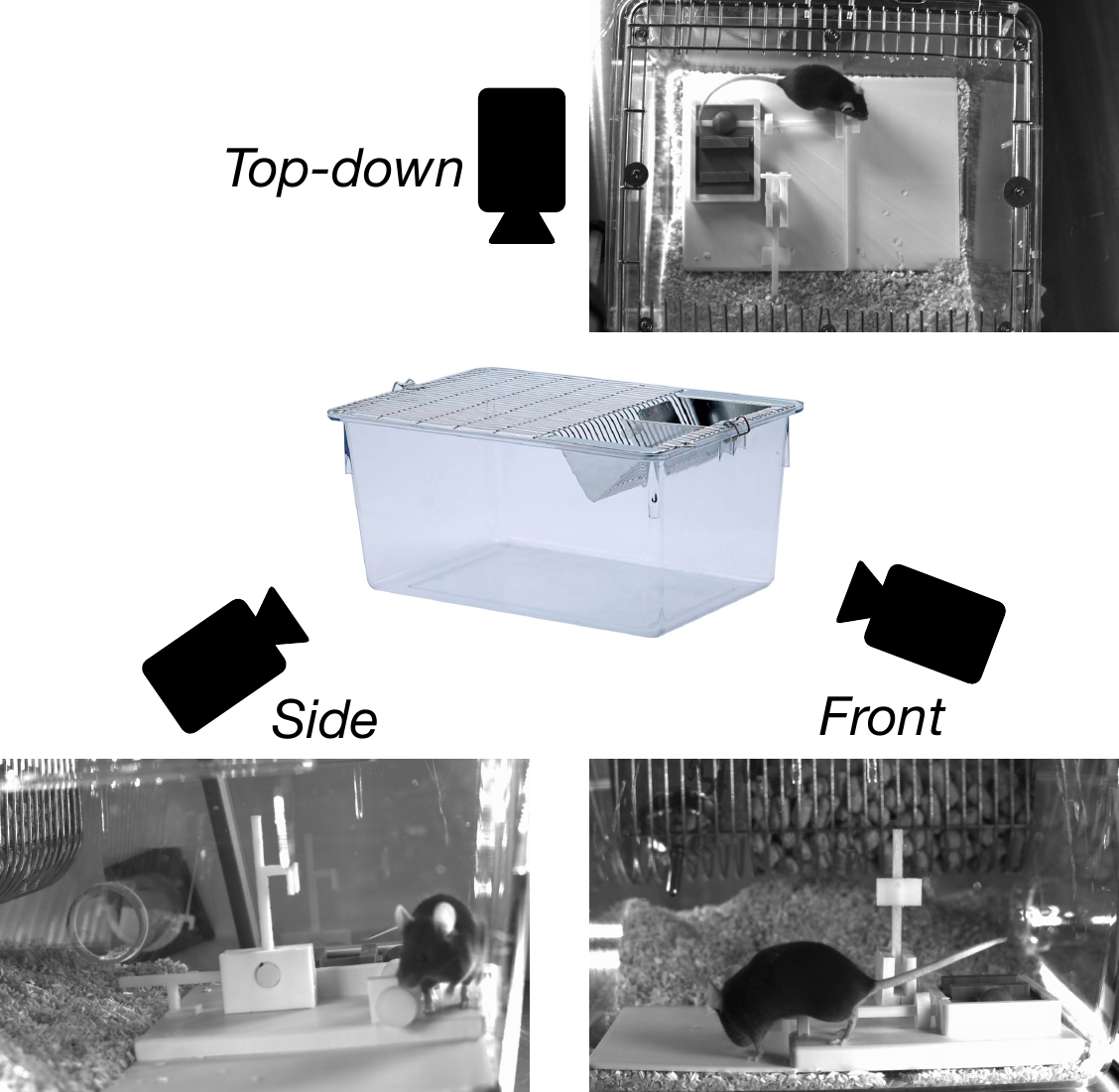}
        \caption{Schematic of the camera setup with named perspectives.}
        \label{fig:camera-setup}
    \end{subfigure}%
    \hspace{\bigskipamount}  
    \begin{subfigure}[b]{.38\linewidth}
        \centering
        \includegraphics[width=\textwidth]{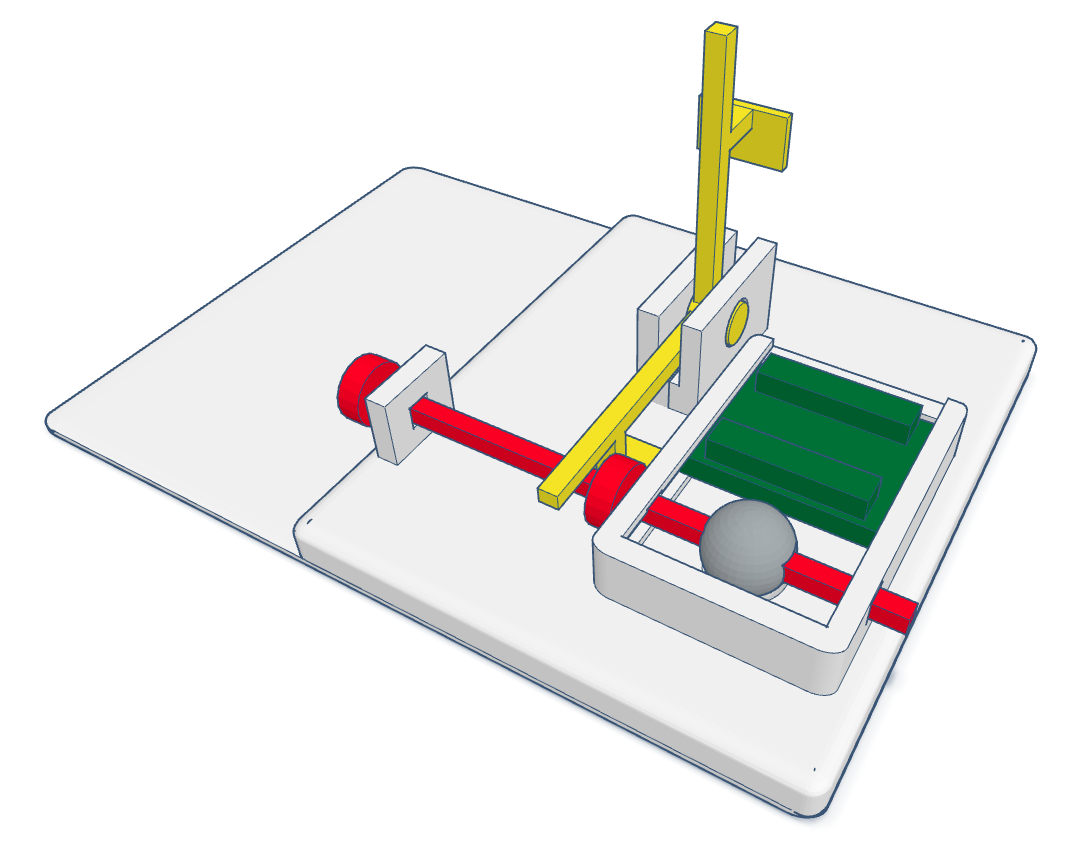}
        \caption{Lockbox of combined mechanisms bait\-ed with a food reward underneath the sliding door.}
        \label{fig:combined-mechanisms}
    \end{subfigure}
    \hfill~
    \begin{subfigure}[b]{\linewidth}
        \medskip
        \centering
        \includegraphics[width=\linewidth]{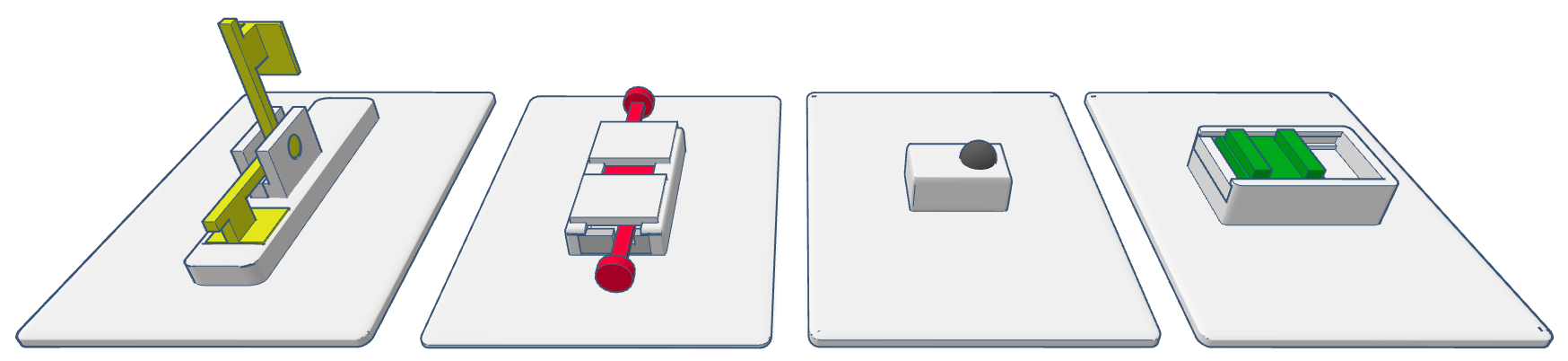}
        \caption{
            Single-mechanism lockboxes, each baited with a food reward.
        }
        \label{fig:single-mechanisms}
    \end{subfigure}
    \caption{
        Camera setup used for recording the videos, as well as lockboxes and their mechanisms: lever (yellow), stick (red), ball (gray), and sliding door (green). 
        Each lockbox is baited with a food reward underneath the (last) mechanism. 
        \cref{apx:opened-lockbox} of our supplementary material provides figures of the unlocked lockboxes.
    }
    \label{fig:setup}
\end{figure}

\cref{fig:camera-setup} depicts the described setup.
All cameras were connected to a computer and controlled by software to synchronize frame capturing.
The mice were presented with five different lockboxes: a combined lockbox~(\cref{fig:combined-mechanisms}) consisting of four interlocked mechanisms, and four simpler lockboxes~(\cref{fig:single-mechanisms}) presenting these mechanisms individually. 
A hidden food reward (oatmeal flake) was used as a bait.
The mice were not subjected to food or water deprivation, but had \textit{ad libitum} access to food pellets (LASvendi, LAS QCDiet, Rod~16, autoclavable) as well as tap water. 
However, the food reward was exclusively provided within the lockboxes. 
To familiarize the mice with the food reward, they were habituated over three consecutive days prior to the lockbox trials by placing eight oat flakes at the location where the lockbox would be placed in the arena.
The freely behaving mice were presented with the combined lockbox for a total of 6 and with the single-mechanism lockboxes 11~trials.
The mice were first presented with the combined lockbox trial followed by 11 trials of a randomized order of each of the single-mechanism lockboxes, followed by another 5 combined lockbox trials.
The videos end shortly after the reward is reached, or if a trial reached the maximum duration of 30~minutes for combined and 15~minutes for single-mechanism lockboxes.
\par 

We manually cut the videos to remove disturbances, e.g., the experimenter's hands switching lockboxes.
Any videos where the lockbox mechanisms were not captured in their entirety were filtered out.
\par 

\subsection{Label Annotation}
\label{ssec:label-annotation}

We provide human annotations for mechanism states, mouse-to-mechanism proximity, as well as both mouse-mechanism and mouse-reward action labels.
To prevent information leakage between labeled and unlabeled data splits, we labeled all videos of two specific mice (mouse numbers 291 and 324) that have a combined total playtime of 15~hours and 25~minutes in 270~videos spanning 90~trials.
This equals about 13\% of our dataset.
\par 

\cref{tbl:ethogram} defines the ethogram used by our nine skilled human raters.
We used these labels as they express trivial truths in order to minimize anthropomorphic biases, that would otherwise distort both experiment evaluations and drawn conclusions.
These biases are especially apparent when using more high-level labels, such as exploring and deliberately manipulating lockbox mechanisms, that strongly depend on subjective human interpretation.
Using more explicit labels not only leads to higher label quality but also lowers the risk of computer vision and machine learning models learning said biases before reintroducing them as noise to any analysis based on their outputs.
\par 

\begin{table}
    \centering
    \begin{tabular}{@{}p{.15\linewidth}p{.775\linewidth}@{}}
        \toprule
        \bfseries LABEL & \multicolumn{1}{c}{\bfseries DEFINITION} \\
        \midrule
        \bfseries Proximity & The mouse's snout is within a distance of 1cm to a specific mechanism. \\[1ex]
        \bfseries Touch & The mouse touches a specific mechanism with one or both of its front paws. \\[1ex]
        \bfseries Bite & The mouse bites into a specific mechanism. \\[1ex]
        \bfseries Unlock & The state of a specific mechanism changes to unlocked. This may make the reward accessible or enabling the next mechanism to be unlocked. State changes may occur without the mouse manipulating a mechanism directly. \\[1ex]
        \bfseries Lock & The state of a specific mechanism changes to locked. This may make the reward inaccessible or preventing the next mechanism from being unlocked. State changes may occur without the direct manipulation of a mechanism. \\[1ex]
        \bfseries Reach reward & The mouse is in first contact with the reward with any of its body parts. \\
        \bottomrule
    \end{tabular}
    \caption{
        Ethogram used for label annotation.
        \cref{apx:example-frame} of our supplementary material provides example frames for the different labels.
    }
    \label{tbl:ethogram}
\end{table}

For annotating the labels, we merged every video triplet (top-down, side, and front perspective) into a combined video.\!\footnote{
    Merging the video triplets into combined videos was necessary as BORIS version 8.27 suffers from a software issue that occurs more frequently when using it with multiple videos opened at once, and that causes to the software to crash only minutes into using it.
    The published dataset does not include the merged videos.
}
All labels have been annotated by randomized pairs of raters with a temporal accuracy of $\pm$100~milliseconds, i.e., $\pm$3~frames, using BORIS~\citep{Friard2016BORIS}.
To annotate any of the videos, it took our raters about 6.2~to 11.5~times longer than their playtime.
This aligns with the factor of about 5~to~10 that is usually reported throughout the available literature.
We account our slightly higher efforts to the multitude of mouse body parts and lockbox mechanisms that needed to be observed at the same time.
\par

\subsection{Dataset Statistics} 
\label{ssec:dataset-statistics}

This section gives an overview over various data and label statistics.
\cref{sssec:playtime-statistics} provides insight on playtime statistics, and
\cref{sssec:label-statistics} on label statistics.
\iftoggle{cvprfinal}{
    We provide further statistics and analyses in our previous work \citep{Hohlbaum2024Enrichment,Boon2024}.
}{}
\par

\subsubsection{Playtime Statistics} 
\label{sssec:playtime-statistics}

Our dataset was recorded from 3~perspectives (top-down, side, and front) and has a total playtime of 117~hours and 52~minutes, i.e., 39~hours and 17~minutes of recorded real time.
It consists of a total of 1629~videos, i.e., 543~trials, with a mean playtime of 4~minutes and 21~seconds.
Any combination of individual mice and lockboxes accounts for 0.3--7.6\% of the data.
Each mouse accounts for 5.3--15.3\% of the data,  while the single-mechanisms lockboxes account for 9.7--14.2\% and the combined lockbox for 52\% of the data.
\par 

\subsubsection{Label Statistics}
\label{sssec:label-statistics}

We provide human-annotated labels for a total playtime of 15~hours and 25~minutes, i.e., 5~hours and 8~minutes of recorded real time.
\cref{ssec:label-annotation} provides details on both the labels and their annotation.
\cref{tbl:action-label-statistics} shows the resulting label distribution.
The uneven label distribution is rooted in the mice behaving freely, and reflects their naturally occurring preference for different actions and mechanisms.
\par 

\begin{table}
    \centering
    \begin{tabular}{@{}r@{}r@{}r@{}r@{}r@{}r@{}}
         \toprule
         & \multicolumn{1}{p{1cm}@{}}{\hfill \bfseries Lever } & \multicolumn{1}{p{1cm}@{}}{\hfill \bfseries Stick} & \multicolumn{1}{p{1cm}@{}}{\hfill \bfseries Ball} & \multicolumn{1}{p{1cm}@{}}{ \hfill \bfseries Door} & \multicolumn{1}{p{1cm}@{}}{\hfill \bfseries Any} \\
         \midrule
         \bfseries Proximity & 15.73 & 19.05 & 13.41 & 18.97 & 55.39 \\
         \bfseries Touch     &  7.06 &  4.07 &  7.00 &  9.32 & 25.50 \\
         \bfseries Bite      &  1.81 &  1.50 &  3.41 &  1.42 & 8.12 \\
         \bottomrule
    \end{tabular}
    \caption{
        Action labels in the labeled videos in percent.
    }
    \label{tbl:action-label-statistics}
\end{table}

\begin{figure}
    \centering\includegraphics[width=\linewidth]{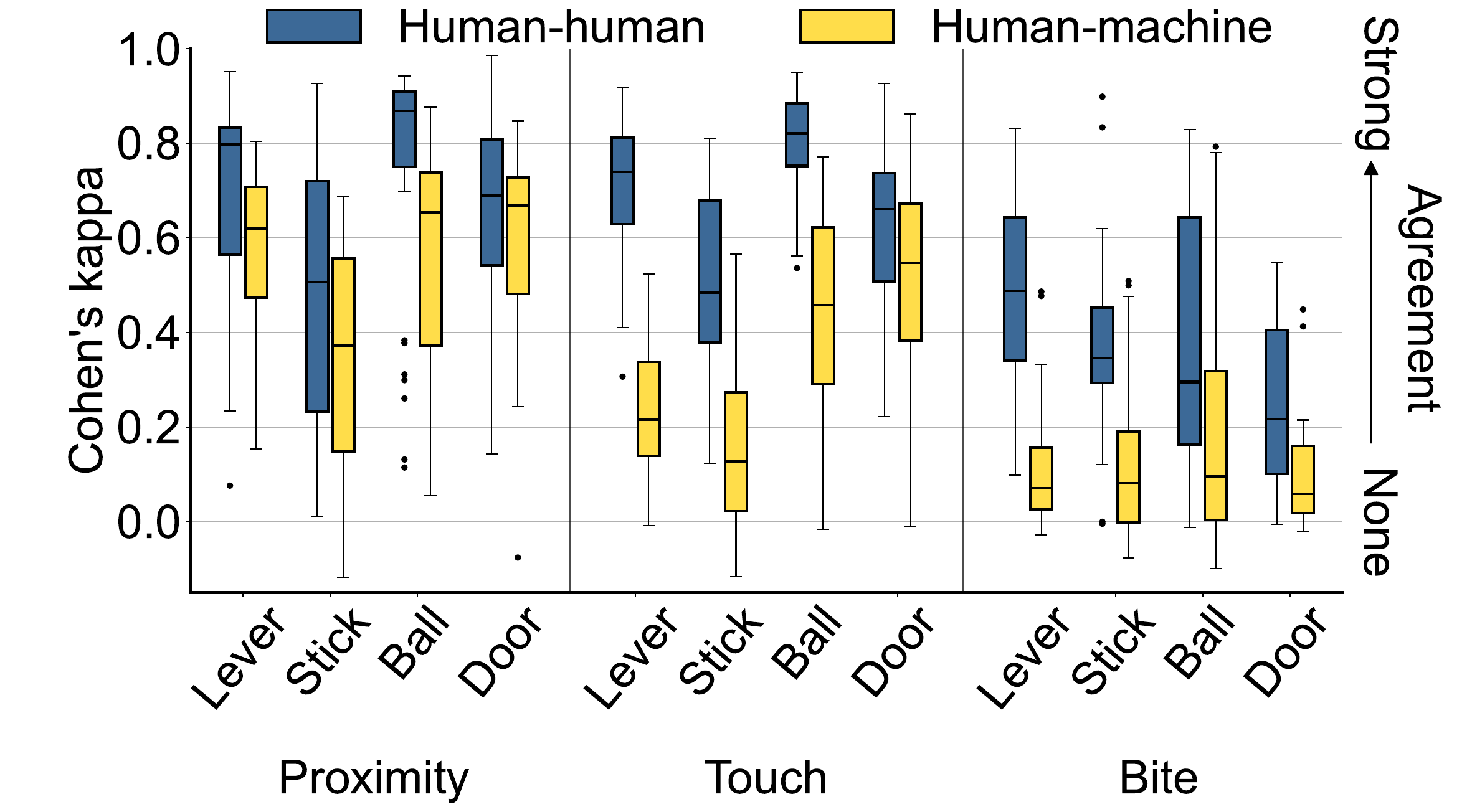}
    \caption{
        Human-human and human-machine inter-rater reliability, i.e., Cohen's kappa coefficients, for different action labels.
    }
    \label{fig:inter-rater_reliability}
\end{figure}

\cref{fig:inter-rater_reliability} shows the inter-rater reliability, i.e., Cohen's kappa coefficients,~\citep{McHugh2012} among human raters.
On average, our human raters annotate most proximity and touch labels with moderate to strong agreement, except for the stick mechanism.
However, they annotate bite labels only with minimal to weak agreement.
We account this to biting being particularly hard to annotate as it rarely is directly visible in the videos.
\cref{fig:inter-rater_reliability} also shows the inter-rater reliability between the human raters and our benchmark method~\citep{Boon2024}, on which we provide details in \cref{apx:benchmark-method} of our supplementary material.
In short, the benchmark method uses 2-dimensional keypoints, extracted with DeepLabCut ~\citep{Mathis2018DLC,NathMathisetal2019DLC}, reconstructs the scene in 3D, and refines the tracks using (extended) Kalman filtering.
These refined 3-dimensional tracks are then used to extract frame-per-frame action labels based on bounding boxes around the mechanisms of interest. 
While it almost reaches human-human reliability for proximity labels, its reliability is outperformed by humans for touch and bite labels.
We account the decreased performance in touch and bite labels to the necessity for increased spatial accuracy in tracking the mechanisms and mouse body parts.
\par

In addition to the inter-rater reliability, we report the F$_1$ scores for all action labels (\cref{tab:f1_scores}) where we use tolerance of $\pm$3 frames that matches the temporal accuracy our human annotators were instructed to adhere to.
As we have mentioned before, annotating actions is a nontrivial task even for humans.
Therefore, we report on both the F$_1$ scores comparing humans against each other, and individual humans against our pipeline, as well as the union of human-annotated labels against our pipeline.
We consider the inter-human F$_1$ scores to be the upper performance limit for any benchmark method.
\par

\begin{table*}[]
    \centering
    \begin{tabular}{@{}l@{~$\times$~}lrrrr|rrrr|rrrr@{}}
        \toprule
        \multicolumn{2}{c}{} & \multicolumn{4}{c|}{\bfseries Proximity} & \multicolumn{4}{|c|}{\bfseries Touch} & \multicolumn{4}{|c@{}}{\bfseries Bite} \\
        \multicolumn{2}{c}{} & \bfseries Lever & \bfseries Stick & \bfseries Ball & \multicolumn{1}{r|}{\bfseries Door} & 
        \bfseries Lever & \bfseries Stick & \bfseries Ball & \multicolumn{1}{r|}{\bfseries Door} & \bfseries Lever & \bfseries Stick & \bfseries Ball & \bfseries Door \\
        \midrule
        \bfseries Human 1 & \bfseries human 2 & 0.85 & 0.70 & 0.88 & 0.83 & 0.76 & 0.63 & 0.88 & 0.72 & 0.62 & 0.32 & 0.58 & 0.17 \\
        \bfseries Human 1 & \bfseries machine & 0.76 & 0.69 & 0.73 & 0.77 & 0.38 & 0.25 & 0.60 & 0.60 & 0.15 & 0.13 & 0.23 & 0.09 \\
        \bfseries Human 2 & \bfseries machine & 0.74 & 0.54 & 0.73 & 0.77 & 0.37 & 0.28 & 0.60 & 0.65 & 0.12 & 0.14 & 0.24 & 0.13 \\
        \bfseries $\cup$ humans & \bfseries machine & 0.77 & 0.64 & 0.74 & 0.80 & 0.40 & 0.29 & 0.63 & 0.67 & 0.19 & 0.19 & 0.30 & 0.18 \\
        \bottomrule
    \end{tabular}
    \caption{F$_1$ scores ($\pm$ 3 frames tolerance) comparing human annotators and our benchmark method. 
    Our benchmark method is compared against both individual humans and the union of their annotated labels.
    The scores comparing humans (top row) can be used as a reference.}
    \label{tab:f1_scores}
\end{table*}

\subsection{Limitations} 
\label{ssec:limitations}

Our dataset has three limitations.
First, since the video recording was pseudo-synchronized through software, the frames of different cameras have been captured with a temporal desynchronization.
We sampled the average asynchronicity to be 1.39~frames with a standard deviation of 1.50~frames.
Since this is lower than the accuracy we annotated our labels with, we do not expect it to cause any major issues.
Second, not all videos share the same exact positioning of the cameras as they have been recorded in several trial periods over the course of months, during which the setup had to be rearranged.
And third, due to insufficient lighting conditions and severe camera dislocation, some trials had to be discarded from the dataset resulting in an imbalanced number of videos per mouse.
\par

\section{Conclusion}
\label{sec:conclusion}

In this work, we presented the---to the best of our knowledge---first available single-agent, multi-perspective video dataset of mice showing complex behavior as they learn to solve mechanical puzzle mechanisms.
These so-called lockboxes consist of either one of four mechanisms or their combination, and are baited with a food reward.
In total, we provide videos with a total playtime, i.e., the number of perspectives multiplied by the real time recorded, of 117~hours and 52~minutes.
\par 

As a benchmark, we provide human-annotated behavior labels for 13\% of our dataset, and report the inter-rater reliability among humans as well as between humans and our benchmark method.
We find that the benchmark method almost reaches human-level performance for proximity labels, but not for touching and biting labels.
\par 

We hope that our dataset will contribute to this advancement by challenging and inspiring others.
\iftoggle{cvprfinal}{
    Our dataset is publicly available at DOI \url{https://doi.org/10.14279/depositonce-23850}
}{
    An anonymized preview of our dataset is available for the reviewers of this manuscript at \url{https://dropbox.com/scl/fo/h7nkai8574h23qfq9m1b2/AP4gNZOpDJJ7z0yGtbWQiOc?rlkey=w36jzxqjkghg0j0xva5zsxy2v}
}
\par 

\iftoggle{cvprfinal}{
    \section*{Acknowledgments}

We thank our encouraged lab assistants Clara Bekemeier, Sophia Meier, Jule Detmers, and Andreas Pauli for their support with cleaning the raw video data and annotating the labels.
Their dedication and hard work were essential to composing the presented dataset. 
\par

This project was funded by the Deutsche Forschungsgemeinschaft (DFG, German Research Foundation) under Germany's Excellence Strategy---EXC 2002/1 ``Science of Intelligence''---project number 390523135.
\par  
}{}
\section*{Ethics Statement}

This research does not involve human subjects, sensitive data, harmful insights, nor methodologies or applications that may raise ethical concerns. 
\par 
 
\iftoggle{cvprfinal}{As reported in~\citep{Hohlbaum2024Enrichment}, animal}{Animal} research was conducted in compliance with the \iftoggle{cvprfinal}{German Animal Welfare Act and Directive \mbox{2010/63/EU}}{national laws and regulations} on the protection of animals used for scientific purposes. 
The experimental procedures and maintenance of the animals were preregistered \iftoggle{cvprfinal}{in the Animal Study Registry (DOI \href{https://doi.org/10.17590/asr.0000237}{\tt 10.17590/asr.0000237})}{} and approved by the \iftoggle{cvprfinal}{Berlin State Authority, Landesamt für Gesundheit und Soziales (permit number \mbox{G0249/19})}{local government}.
\par 

The authors declare that they have no conflicts of interest. 
No sponsorships influenced this research.
\par
{
    \small
    \bibliographystyle{ieeenat_fullname}
    \bibliography{references}
}

\clearpage
\setcounter{page}{1}
\maketitlesupplementary

\section{Lockboxes with Unlocked Mechanisms}
\label{apx:opened-lockbox}

\begin{figure}
    \centering
    \begin{subfigure}{\linewidth}
        \centering
        \includegraphics[width=\linewidth]{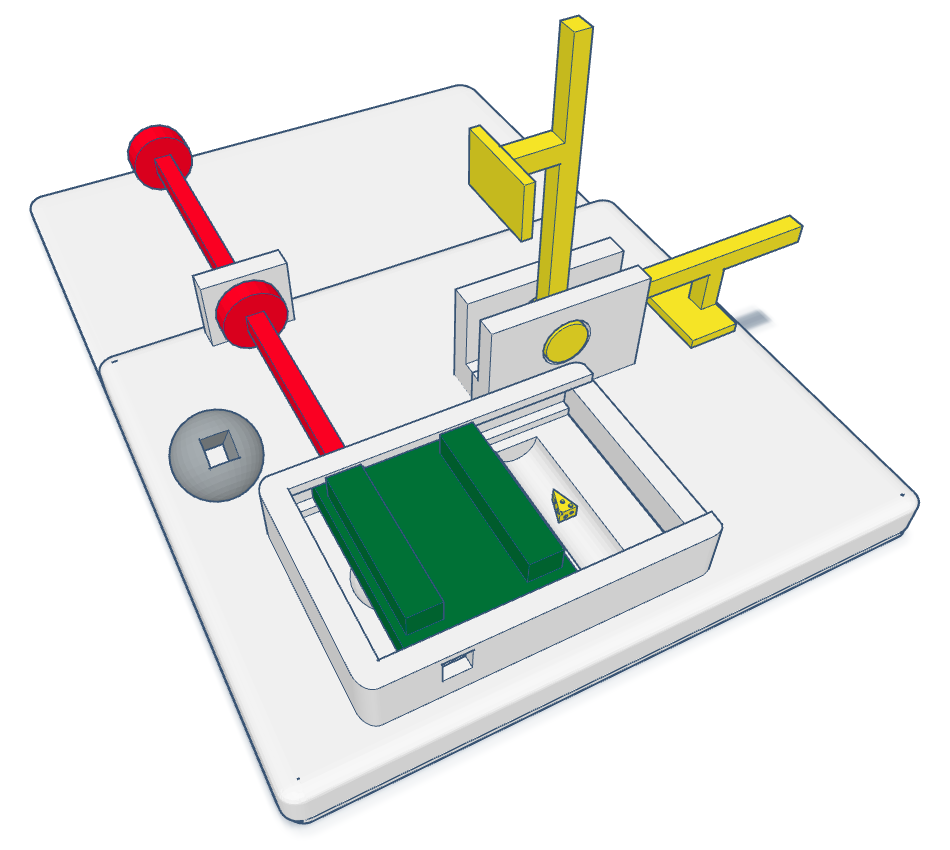}
        \caption{
            Unlocked lockbox of combined mechanisms baited with a symbolized food reward underneath the sliding door.
        }
        \label{fig:combined-mechanisms-open}
    \end{subfigure}
    \\ \bigskip
    \begin{subfigure}{\linewidth}
        \medskip
        \centering
        \includegraphics[width=\linewidth]{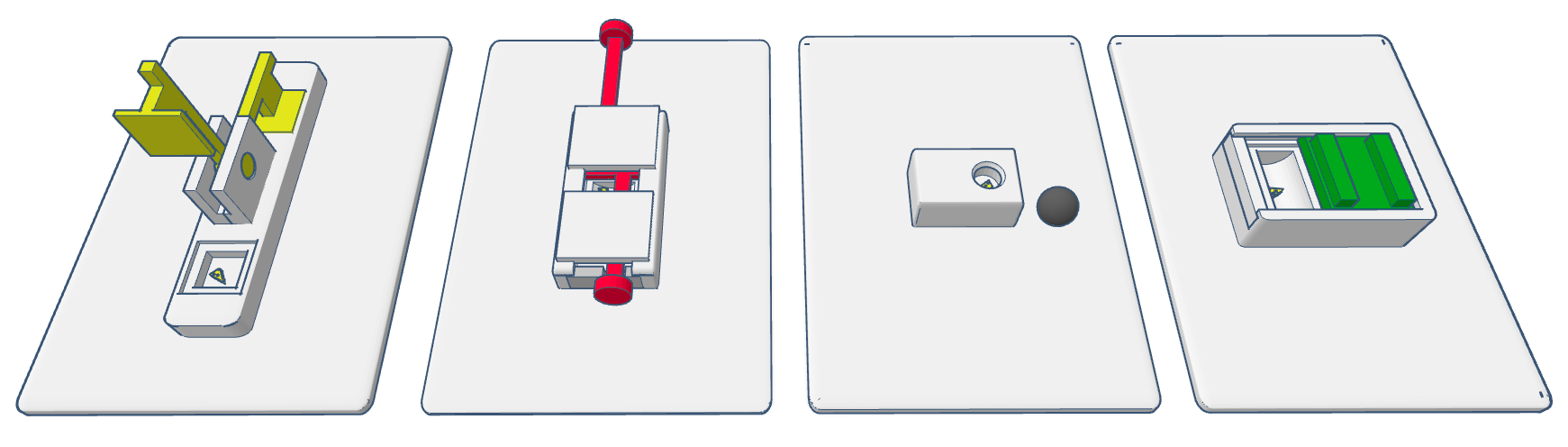}
        \caption{
            Unlocked single-mechanism lockboxes baited with a symbolized food reward underneath each mechanisms.
        }
        \label{fig:single-mechanisms-open}
    \end{subfigure}
    \caption{
        Unlocked lockboxes and their mechanisms: lever (yellow), stick (red), ball (gray), and sliding door (green). 
        This depiction contains symbolized food baits. 
    }
    \label{fig:setup-open}
\end{figure}

\cref{fig:setup-open} shows the opened lockboxes with symbolized food baits; see \cref{fig:combined-mechanisms,fig:single-mechanisms} for reference. 
\par 

\section{Example Frames for Labels}
\label{apx:example-frame}

\begin{figure*}[p]
    \centering
    \begin{subfigure}{\linewidth}
        \includegraphics[width=\linewidth]{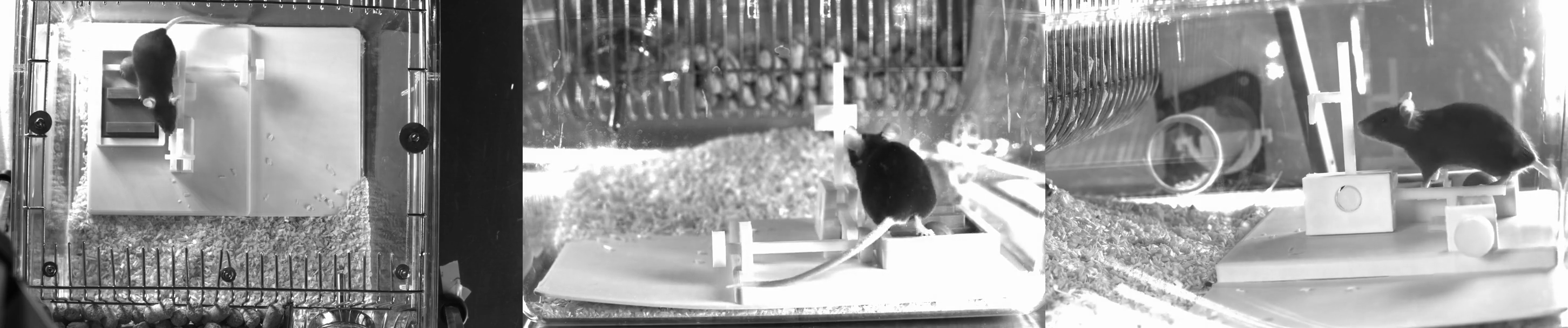}
        \caption{Frame example with mouse in proximity to lever and touching the sliding door while all mechanisms are locked.}
    \end{subfigure}
    \\ \bigskip
    \begin{subfigure}{\linewidth}
        \includegraphics[width=\linewidth]{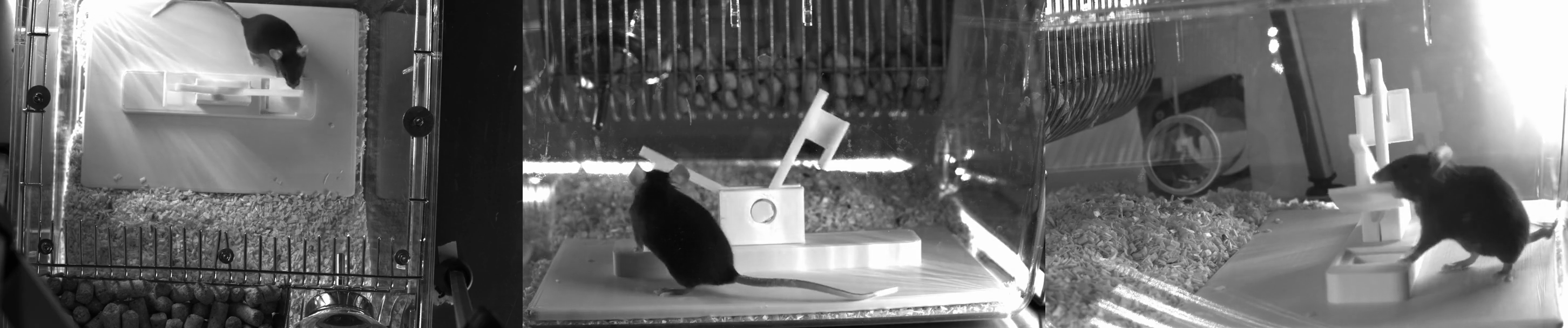}
        \caption{Frame example with mouse in proximity to and biting the lever while the mechanism is unlocked.}
    \end{subfigure}
    \\ \bigskip
    \begin{subfigure}{\linewidth}
        \includegraphics[width=\linewidth]{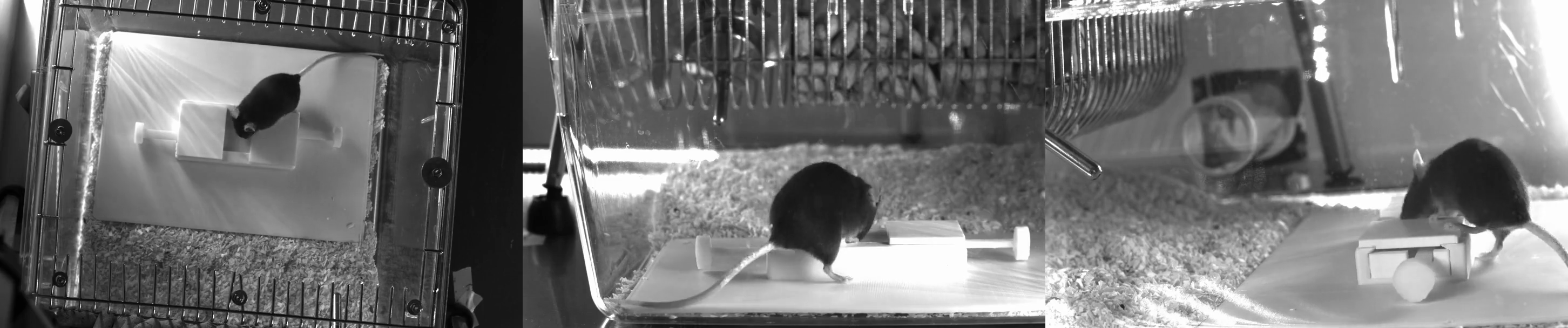}
        \caption{Frame example with mouse in proximity to the stick while the mechanism is locked.}
    \end{subfigure}
    \\ \bigskip
    \begin{subfigure}{\linewidth}
        \includegraphics[width=\linewidth]{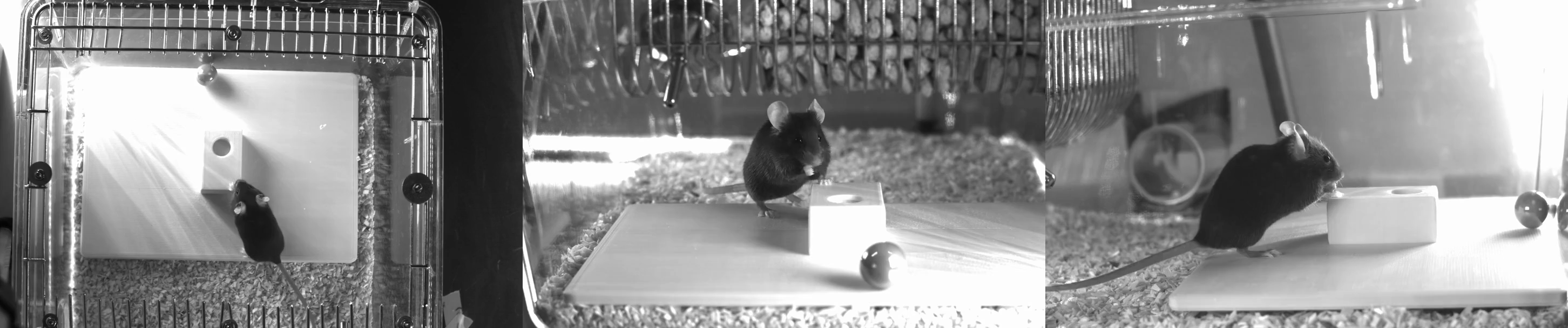}
        \caption{Frame example with no action label active while the ball mechanism is unlocked. }
    \end{subfigure}
    \\ \bigskip
    \begin{subfigure}{\linewidth}
        \includegraphics[width=\linewidth]{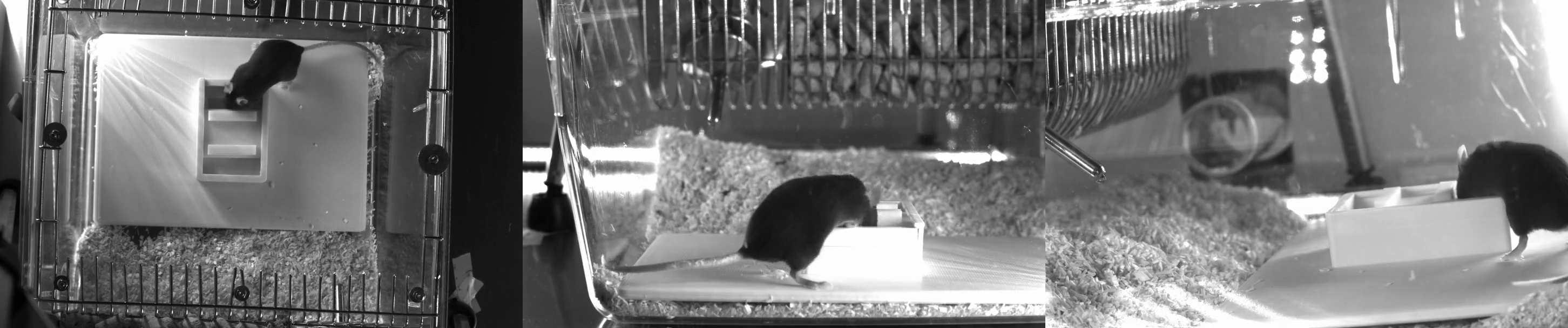}
        \caption{Frame example with mouse in proximity to the sliding door while the mechanism is unlocked.}
    \end{subfigure}
    \caption{Example frames from labeled videos showing mice performing different actions.}
    \label{fig:example-frames}
\end{figure*}

Figure~\ref{fig:example-frames} shows a selection of examples for our different label classes.
\par 

\section{Benchmark Method}
\label{apx:benchmark-method}

Our benchmark experiments are based on the pose-tracking approach used by \citet{Boon2024}. 
The method consists of three steps: the use of DeepLabCut (DLC) for 2-dimensional pose tracking, 3-dimensional reconstruction and the refinement of keypoint data using (Extended) Kalman filtering, and the detection of action labels. 
A high-level description of steps is given below.
\par

First, 2-dimensional poses of the mice and lockbox mechanisms are extracted from the videos on a frame-level by learning DLC models under supervision.
We learn one DLC model to locate keypoints of mice, and two that locate keypoints of lockbox mechanisms---one for the single-mechanism lockboxes, and one for the lockbox combining them---using default parameters~\citep{Mathis2018DLC,NathMathisetal2019DLC}.
Next, the scene is reconstructed by utilizing the known 3-dimensional locations of the lockbox mechanisms given by their CAD models.
We linearly map the known 3-dimensional locations onto the corresponding triplets of 2-dimensional keypoints using the RANSAC algorithm and construct a triangulation matrix for each trial. 
For trials in which the lockbox does not have a well-defined third dimension (i.e., single stick, ball, and door), the lever trial of that mouse (and day) provides reference triangulation data instead.
Potential rotations due to the lockbox not being placed in the same orientation as the lever reference are accounted for by rotating the 3-dimensional reference coordinates from the CAD model in accordance to the observed rotation in the xy-plane from the top-down view camera.
\par
The triangulation matrices for each trial are used as observation matrices for (Extended) Kalman filters to refine the observed triplets of 2-dimensional keypoints into a common 3-dimensional space.
The head and the tail of the mouse are inferred using a skeletal model, while the keypoints of the mechanisms and the paws of the mouse are inferred as single keypoints.
\par
Finally, the interactions of the mice with the lockbox mechanisms are detected based on the 3-dimensional poses of the mouse and predefined bounding boxes spanned by the 3-dimensional keypoint locations.
For the proximity labels, the snout of mouse is used to detect the actions: each frame in which the snout of the mouse is inside of a bounding box defined around each lockbox mechanism, the corresponding action label (e.g., proximity lever) is detected.
Biting is detected using the mouth of the mouse, which location is computed from the rigid body model of the mouse head.
The touch labels are detected using the locations of the front paws.
Note that the bite and touch labels have different predefined bounding boxes than the proximity labels, as these actions have a finer level of granularity than proximity labels.
As a last processing step, each action label is filtered by a weighted moving average filter with a Gaussian shape (with a window size of 30 frames), to remove faulty single-frame detections from the data.

\end{document}